\documentclass{article}

\usepackage{microtype}
\usepackage{graphicx}
\usepackage{subfigure}
\usepackage{booktabs} % for professional tables
\usepackage{algpseudocode}
\usepackage{algorithmicx}

\usepackage[toc,page]{appendix}

\usepackage{blindtext}
\usepackage{hyperref}

\usepackage{icml2021}

\makeatletter
\newenvironment{megaalgorithm}[1][H]{%
    \renewcommand{\ALG@name}{Procedure}% Update algorithm name
   \begin{algorithm}[#1]%
  }{\end{algorithm}}
\makeatother

\icmltitlerunning{Improved Binary Forward Exploration: Learning Rate Optimization}

\begin{document}

\twocolumn[
\icmltitle{Improved Binary Forward Exploration: \\ 
Learning Rate Scheduling Method for Stochastic Optimization}



\icmlsetsymbol{equal}{*}

\begin{icmlauthorlist}
\icmlauthor{Xin Cao}{equal, to, goo}
%\icmlauthor{??}{equal,to,goo}
%\icmlauthor{??}{goo}
%\icmlauthor{??}{ed}
%\icmlauthor{??}{to}
%\icmlauthor{??}{ed,to,goo}
%\icmlauthor{??}{goo}
%\icmlauthor{??}{ed}
%\icmlauthor{??}{ed}
%\icmlauthor{??}{to,goo}
%\icmlauthor{??}{ed}
%\icmlauthor{??}{ed}
\end{icmlauthorlist}

\icmlaffiliation{to}{CU Boulder, Boulder, CO, USA}
\icmlaffiliation{goo}{Boulder, CO, USA}
\icmlaffiliation{ed}{??, ??, ??, ??}

\icmlcorrespondingauthor{Xin Cao}{xincao.tech@gmail.com}

\icmlkeywords{Machine Learning, ICML}

\vskip 0.3in
]



\printAffiliationsAndNotice{\icmlEqualContribution} % otherwise use the standard text.

\begin{abstract}
A new gradient-based optimization approach by automatically scheduling the learning rate has been proposed recently, which is called Binary Forward Exploration (BFE). The Adaptive version of BFE has also been discussed thereafter. In this paper, the improved algorithms based on them will be investigated, in order to optimize the efficiency and robustness of the new methodology. This improved approach provides a new perspective to scheduling the update of learning rate and will be compared with the stochastic gradient descent (SGD) algorithm with momentum or Nesterov momentum and the most successful adaptive learning rate algorithm e.g. Adam. The goal of this method does not aim to beat others but provide a different viewpoint to optimize the gradient descent process. This approach combines the advantages of the first-order and second-order optimizations in the aspects of speed and efficiency.
\end{abstract}

\section{Introduction}
\label{submission}

In the Statistics and Machine Learning areas, gradient descent algorithm is widely used to optimize the model parameters via searching the local minimum of the objective function or loss function. The algorithm and its variants play very important roles in machine learning, deep learning and reinforcement learning. The reason that the first-order gradient-based optimization methods are so popular is mainly because of their low computational complexity and their simplicity of implementation, although the second-order methods usually have faster convergence speed. The first-order algorithm has two categories: non-adaptive and adaptive. The stochastic gradient descent (SGD), which enables all parameters sharing the same updated learning rate, is one of the most famous non-adaptive approach. Some extensions of SGD such as momentum and Nesterov momentum via tracking historic gradients during the optimization \cite{nesterov1983method}, provide higher convergence speed and stronger robustness. Meanwhile, the recently developed adaptive methods such as AdaGrad \cite{duchi2011adaptive}, Adadelta \cite{zeiler2012adadelta}, RMSProp \cite{tieleman2012lecture}, and Adam \cite{kingma2014adam}, update the learning rate per-parameter, and make the convergence faster compared to the non-adaptive methods. However, in practice, the traditional SGD method usually approach better optima in many deep learning tasks, which raises questions e.g. What is the relationship between the learning rate updates and the loss landscape of such tasks? In my previous paper, a new method that automatically scheduling the learning rate by detecting the forward information of the loss landscape has been introduced, which is called Binary Forward Exploration (BFE) algorithm \cite{2022arXiv220702763C}. In this paper, a modification to the BFE algorithm will be discussed in detail, in order to improve the optimization efficiency and reduce the computational cost. It is proposed that this type of optimization approach can be applied to the interdisciplinary study with physics problems in the future, such as some space physics studies based on analyzing the spacecraft data of different space missions \cite{cao2022machine, liu2012dipole}, and \cite{chu2021dayside, chu2021electrostatic, liuzzo2021investigating}. It is also expected that more numerical methods sourced from computational physics is probably able to be transferred and applied to the machine learning problems.

\section{Improved Binary Forward Exploration Algorithm}

\subsection{Algorithms}

The improved Binary Forward Exploration Method was created to compare the loss values in the forward direction via computing an original learning rate and a binary reduction or amplification of the learning rate, which derivates from the numerical modeling work \cite{Cao2013Trajectorymethodof3Dtestparticles} and modified based on the original BFE approach \cite{2022arXiv220702763C} .

The suggested method's pseudo-code is shown in algorithm 1, where $\frac{\partial f(\theta)}{\partial \theta_{t}}$ is the gradient of the loss function at $\theta_{t}$, and $\frac{\partial f(\theta)}{\partial \theta^{+}_{t}}$ represents the gradient of the loss function at $\theta^{+}_{t}$. The gradient $\frac{\partial f(\theta)}{\partial \theta_{t}}$ represents to $\nabla f(\theta_{t})$ in this article, where $\theta_{t}$ denotes the parameter vectors or tensors at time-step t, and $f(\theta)$ is the loss function with respect to the parameter.

\begin{algorithm}[H]
   \caption{Improved BFE, the proposed algorithm in non-adaptive learning rate automation for stochastic optimization. More details are described in the text. The Loss Function can be written as a function w.r.t parameters, e.g. $Loss=f(\theta)$. Default value of the error limit ratio for Binary Detection Learning Rate is $\epsilon = 0.001$, or one thousandth.}
   \label{alg:example1}
\begin{algorithmic}
   
   \State Initialize learning rate $\eta$ (e.g. 0.001)
   \State Initialize $\epsilon_v$ (e.g. 0.001)
   % \STATE \textcolor{orange}{Initialize} $\epsilon_v$ (e.g. 0.001)
   \State Initialize $\epsilon_c > \epsilon_v$
   \State Initialize parameter vector $\theta_0$
   \State Initialize time-step $t=0$
   \While{$\theta_t$ not converged}
     \State $t=t+1$
     \If{$\epsilon_c \geq \epsilon_v$}
       \While{$\epsilon_c \geq \epsilon_v$}
         \State $\theta^{*}_{t} = \theta_t - \eta \frac{\partial f(\theta)}{\partial \theta_{t}}$
         \State $\theta^{+}_{t} = \theta_t - \frac{\eta}{2} \frac{\partial f(\theta)}{\partial \theta_{t}}$
         \State $\theta^{'}_{t} = \theta^{+}_{t} - \frac{\eta}{2} \frac{\partial f(\theta)}{\partial \theta^{+}_{t}}$
         \State $Loss1 = [f(\theta)]_{\theta^{*}_{t}}$ $ $ $ $ (loss value at $\theta^{*}_{t}$)
         \State $Loss2 = [f(\theta)]_{\theta^{'}_{t}}$ $ $ $ $ (loss value at $\theta^{'}_{t}$)
         \State $\epsilon_c = |Loss2-Loss1|$
         \State $\epsilon_v= 0.5 \cdot (|Loss2|+|Loss1|) \cdot \epsilon$
         \State \ \ \ \ \ \ \ \ \  or $min(|Loss2|\cdot \epsilon, |Loss1|\cdot \epsilon)$
         \State \ \ \ \ \ \ \ \ \  or any other predefined factor
         \State \ \ \ \ \ \ \ \ \  or functions, e.g. decay with timestep 
         \State \ \ \ \ \ \ \ \ \  or epochs
         \State $\eta = \frac{\eta}{2}$
       \EndWhile
       \State $\eta = 2\eta$
       \State $\theta_{t} = \theta^{*}_{t}$
       \State (Procedure updating $\epsilon_c$ and $\epsilon_v$, e.g. as below)
       \State $\theta^{*}_{t} = \theta_t - \eta \frac{\partial f(\theta)}{\partial \theta_{t}}$
       \State $\theta^{+}_{t} = \theta_t - \frac{\eta}{2} \frac{\partial f(\theta)}{\partial \theta_{t}}$
       \State $\theta^{'}_{t} = \theta^{+}_{t} - \frac{\eta}{2} \frac{\partial f(\theta)}{\partial \theta^{+}_{t}}$
       \State $Loss1 = [f(\theta)]_{\theta^{*}_{t}}$ $ $ $ $ (loss value at $\theta^{*}_{t}$)
       \State $Loss2 = [f(\theta)]_{\theta^{'}_{t}}$ $ $ $ $ (loss value at $\theta^{'}_{t}$)
       \State $\epsilon_c = |Loss2-Loss1|$
       \State $\epsilon_v= 0.5 \cdot (|Loss2|+|Loss1|) \cdot \epsilon$
       \State \ \ \ \ \ \ \ \ \  or $min(|Loss2|\cdot \epsilon, |Loss1|\cdot \epsilon)$
       \State \ \ \ \ \ \ \ \ \  or any other predefined factor
       \State \ \ \ \ \ \ \ \ \  or functions, e.g. decay with timestep or
       \State \ \ \ \ \ \ \ \ \  epochs
     \Else
       \While{$\epsilon_c < \epsilon_v$}
         \State $\theta^{+}_{t} = \theta_t - \eta \frac{\partial f(\theta)}{\partial \theta_{t}}$

         \algstore{}
\end{algorithmic}
\end{algorithm}

\begin{algorithm}[H]
%\ContinuedFloat
\begin{algorithmic}
       \algrestore{}
       \State $\theta^{'}_{t} = \theta^{+}_t - \eta \frac{\partial f(\theta)}{\partial \theta^{+}_{t}}$
       \State $\theta^{*}_{t} = \theta_{t} - 2\eta \frac{\partial f(\theta)}{\partial \theta_{t}}$
       \State $Loss1 = [f(\theta)]_{\theta^{'}_{t}}$ $ $ $ $ (loss value at $\theta^{'}_{t}$)
       \State $Loss2 = [f(\theta)]_{\theta^{*}_{t}}$ $ $ $ $ (loss value at $\theta^{*}_{t}$)
       \State $\epsilon_c = |Loss2-Loss1|$
       \State $\epsilon_v= 0.5 \cdot (|Loss2|+|Loss1|) \cdot \epsilon$
       \State \ \ \ \ \ \ \ \ \  or $min(|Loss2|\cdot \epsilon, |Loss1|\cdot \epsilon)$
       \State \ \ \ \ \ \ \ \ \  or any other factor or functions
         \State $\eta = 2\eta$
       \EndWhile
       %\algstore{}
       \State $\eta = \frac{\eta}{2}$
       \State $\theta_{t} = \theta^{+}_{t}$
       \State (Procedure updating $\epsilon_c$ and $\epsilon_v$, e.g. as below)
       \State $\theta^{+}_{t} = \theta_t - \eta \frac{\partial f(\theta)}{\partial \theta_{t}}$
       \State $\theta^{'}_{t} = \theta^{+}_t - \eta \frac{\partial f(\theta)}{\partial \theta^{+}_{t}}$
       \State $\theta^{*}_{t} = \theta_{t} - 2\eta \frac{\partial f(\theta)}{\partial \theta_{t}}$
       \State $Loss1 = [f(\theta)]_{\theta^{'}_{t}}$ $ $ $ $ (loss value at $\theta^{'}_{t}$)
       \State $Loss2 = [f(\theta)]_{\theta^{*}_{t}}$ $ $ $ $ (loss value at $\theta^{*}_{t}$)
       \State $\epsilon_c = |Loss2-Loss1|$
       \State $\epsilon_v= 0.5 \cdot (|Loss2|+|Loss1|) \cdot \epsilon$
       \State \ \ \ \ \ \ \ \ \  or $min(|Loss2|\cdot \epsilon, |Loss1|\cdot \epsilon)$
       \State \ \ \ \ \ \ \ \ \  or any other factor or functions
     \EndIf
   \EndWhile \\
   \Return $\theta_{t}$ (Resulting Optimized Parameters)
   
\end{algorithmic}
\end{algorithm}

At each time-step, this approach compares the values of $\epsilon_{comp}$ and $\epsilon_{val}$ ($\epsilon_{c}$ and $\epsilon_{v}$ for short in the algorithms' pseudo-code), where $\epsilon_{comp}$ is the relative error or inaccuracy of a one-time updated loss function, and $\epsilon_{val}$ is defined as the threshold value that determine if an optimal learning rate is obtained and thus the searching iteration can be terminated. Meanwhile, $\epsilon_{val}$ can be also defined as a decay function with the number of epochs, for instance, $\epsilon_{val} = 0.5 \cdot (|Loss2|+|Loss1|) \cdot \epsilon /(t+t_{decay})$, where $t_{decay}$ is a predefined number of the time-step or epochs (e.g. 100) when the decay starts. Furthermore, besides the binary shrink or amplification of the step size or learning rate, the Binary Forward Exploration can be also extended to the Multiple Forward Exploration, which will be displayed in the Appendix.

\subsection{Improved BFE’s update rule:}

As the original BFE algorithm which \cite{2022arXiv220702763C} has discussed, the absolute value of the subtraction between Loss 1 (LS1) and Loss 2 (LS2) at the current time-step can be used to determine $\epsilon_{comp}$, as shown in algorithm 1. In the "while $\epsilon_{comp} \geq \epsilon_{val}$ do" loop (Zoom-in part),  Loss1 represents the loss calculated at $\theta^{*}_{t}$,  where $\theta^{*}_{t}$ is updated based on $\theta_{t}$  via the gradient descent method combined with the current learning rate. Similarly, Loss2, is calculated as follows: first, update the parameter through half of current learning rate and the gradient at $\theta_{t}$ to compute $\theta^{+}_{t}$, and then update it again via the same learning rate and the gradient at the new position $\theta^{+}_{t}$ to compute $\theta^{'}_{t}$. In the “while $\epsilon_{comp}$ < $\epsilon_{val}$ do” loop (Zoom-out part), Loss1 is specifically defined in this way: first, update the parameters via current learning rate and the gradient at $\theta_{t}$ to compute  $\theta^{+}_{t}$, and then repeat the same way to update it utilizing half of current learning rate and the gradient at $\theta^{+}_{t}$ to compute $\theta^{'}_{t}$. Loss2 is defined as the value of the loss function at $\theta^{*}_{t}$, where $\theta^{*}_{t}$ is updated based on $\theta_{t}$ via the gradient descent combined with twice of current learning rate. Using the same way as described in \cite{2022arXiv220702763C}, the learning rate can be updated during each time-step’s iteration by comparing Loss1 and Loss2 in the zoom-in or zoom-out parts as mentioned above.

The main difference from the original BFE algorithm is marked as the bold pseudo-code as Algorithm 1 shows. The learning rate is doubled after the zoom-in iteration for each time-step and thus the parameters are correspondingly updated, which can speed up the updating process without losing accuracy. Afterwards, $\epsilon_{comp}$ and $\epsilon_{val}$ are then updated as well at the end of zoom-in part. Similarly, $\epsilon_{comp}$ and $\epsilon_{val}$ are also updated at the end of zoom-out part. Such an additional update can make the algorithm detect the variation of the loss landscape in the forward direction more accurately, which can also speed up the convergence of the parameters.

\begin{algorithm}[H]
   \caption{Improved BFE (zoom-in part only), the proposed algorithm in non-adaptive learning rate automation for stochastic optimization. More details are described in the text. The Loss Function can be written as a function w.r.t parameters, e.g. $Loss=f(\theta)$. Default setting for error limit ratio for Binary Detection Learning Rate is $\epsilon = 0.001$, indicating one thousandth.}
   \label{alg:example2}

\begin{algorithmic}
   
   \State Initialize learning rate $\eta = \eta_0$ (e.g. 0.001)
   \State Initialize $\epsilon_v$ (e.g. 0.001)
   % \STATE \textcolor{orange}{Initialize} $\epsilon_v$ (e.g. 0.001)
   \State Initialize $\epsilon_c > \epsilon_v$
   \State Initialize parameter vector $\theta_0$
   \State Initialize time-step $t=0$
   \While{$\theta_t$ not converged}
     \State $t=t+1$
     \State $\eta = \eta_0$
     \State $\theta^{*}_{t} = \theta_t - \eta \frac{\partial f(\theta)}{\partial \theta_{t}}$
     \State $\theta^{+}_{t} = \theta_t - \frac{\eta}{2} \frac{\partial f(\theta)}{\partial \theta_{t}}$
     \State $\theta^{'}_{t} = \theta^{+}_{t} - \frac{\eta}{2} \frac{\partial f(\theta)}{\partial \theta^{+}_{t}}$
     \State $Loss1 = [f(\theta)]_{\theta^{*}_{t}}$ $ $ $ $ (loss value at $\theta^{*}_{t}$)
     \State $Loss2 = [f(\theta)]_{\theta^{'}_{t}}$ $ $ $ $ (loss value at $\theta^{'}_{t}$)
     \State $\epsilon_c = |Loss2-Loss1|$

         \algstore{}
\end{algorithmic}

\end{algorithm}

\begin{algorithm}[H]
\begin{algorithmic}
       \algrestore{}
       \State $\epsilon_v= 0.5 \cdot (|Loss2|+|Loss1|) \cdot \epsilon$
       \State \ \ \ \ \ \ \ \ \  or $min(|Loss2|\cdot \epsilon, |Loss1|\cdot \epsilon)$
       \State \ \ \ \ \ \ \ \ \  or any other predefined factor
       \State \ \ \ \ \ \ \ \ \  or functions, e.g. decay with timestep or
       \ \ \ \ \ \ \ \ \  epochs
       \While{$\epsilon_c \geq \epsilon_v$}
         \State $\theta^{*}_{t} = \theta_t - \eta \frac{\partial f(\theta)}{\partial \theta_{t}}$
         \State $\theta^{+}_{t} = \theta_t - \frac{\eta}{2} \frac{\partial f(\theta)}{\partial \theta_{t}}$
         \State $\theta^{'}_{t} = \theta^{+}_{t} - \frac{\eta}{2} \frac{\partial f(\theta)}{\partial \theta^{+}_{t}}$
         \State $Loss1 = [f(\theta)]_{\theta^{*}_{t}}$ $ $ $ $ (loss value at $\theta^{*}_{t}$)
         \State $Loss2 = [f(\theta)]_{\theta^{'}_{t}}$ $ $ $ $ (loss value at $\theta^{'}_{t}$)
         \State $\epsilon_c = |Loss2-Loss1|$
         \State $\epsilon_v= 0.5 \cdot (|Loss2|+|Loss1|) \cdot \epsilon$
         \State \ \ \ \ \ \ \ \ \  or $min(|Loss2|\cdot \epsilon, |Loss1|\cdot \epsilon)$
         \State \ \ \ \ \ \ \ \ \  or any other predefined factor
         \State \ \ \ \ \ \ \ \ \  or functions, e.g. decay with time-step
         \State $\epsilon_c = |Loss2-Loss1|$
         \State $\epsilon_v= 0.5 \cdot (|Loss2|+|Loss1|) \cdot \epsilon$
         \State \ \ \ \ \ \ \ \ \  or $min(|Loss2|\cdot \epsilon, |Loss1|\cdot \epsilon)$
         \State \ \ \ \ \ \ \ \ \  or any other predefined factor
         \State \ \ \ \ \ \ \ \ \  or functions, e.g. decay with timestep or
         \State \ \ \ \ \ \ \ \ \  epochs
         \State $\eta = \frac{\eta}{2}$
       \EndWhile

       \State $\eta = 2\eta$
       \State $\theta_{t} = \theta^{*}_{t}$
   \EndWhile \\
   \Return $\theta_{t}$ (Resulting Optimized Parameters)
       
\end{algorithmic}

\end{algorithm}

The zoom-in only version of improved BFE algorithm is shown as Algorithm 2, during the iteration of each time-step, the learning rate is re-set to be the pre-defined value every time before the execution of the zoom-in process, and the Loss1 and Loss2 will be calculated and so are the $\epsilon_{comp}$ and $\epsilon_{val}$. The zoom-in process will be executed to determine which learning rate is optimal for the current time-step via calculating the forward gradients and the corresponding losses. At the end of the zoom-in part, the learning rate is doubled and the parameters are updated based on it, as discussed in Algorithm 1. The algorithm is not terminated until the parameters approach optima.

\section{Experiments}

To investigate the improved method's performance, we utilized the same linear regression model as the model used in \cite{2022arXiv220702763C}. The corresponding loss function is a quadratic function form, which is efficient to measure the behavior of convergence and meanwhile satisfies the characteristics of a local minimum, \cite{sutskever2013importance}, \cite{o2015adaptive}, \cite{lucas2018aggregated}. A more complex distribution of the loss function in a high-dimensional parameter space by using neural network models and different datasets such as MNIST /cite{lecun1998gradient}, CIFAR-10, CIFAR-100 \cite{krizhevsky2009learning}and Mini-ImageNet \cite{vinyals2016matching} will be studied in the future work, which is though beyond the scope of this work.

\begin{figure}[ht]
\vskip 0.2in
\begin{center}
\centerline{\includegraphics[width=\columnwidth]{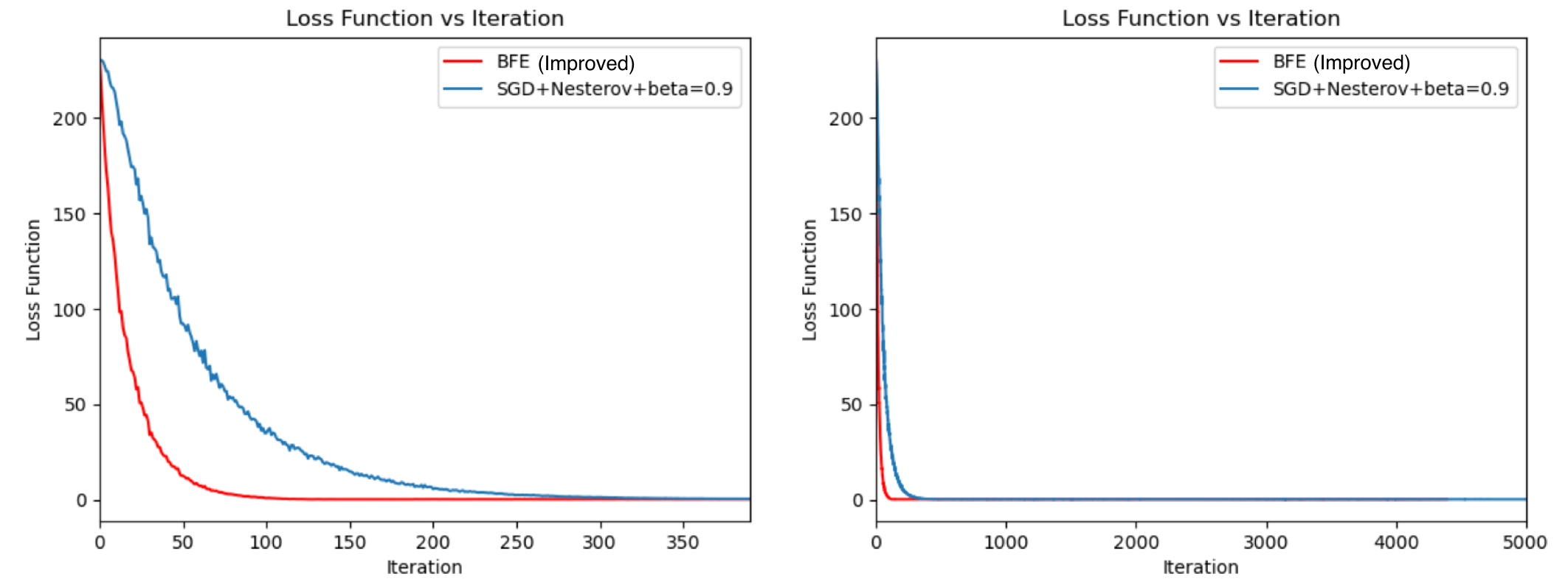}}
\caption{The loss decrease pattern with the increasing iteration over the dataset. The loss decrease by using the BFE optimizer (red curve) and the black represents that by using the mini-batch stochastic gradient descent algorithm (with Nesterov terms of beta = 0.9), which is in contrast to Fig. 3 in \cite{2022arXiv220702763C}}
\label{icml-historical1}
\end{center}
\vskip -0.2in
\end{figure}

The details of the regression model were described at \cite{2022arXiv220702763C}. Figure 1 shows that the improved BFE algorithm enables the loss converge much faster than another classic non-adaptive algorithm: the mini-batch gradient descent method. The left and right panel respectively shows the loss variation from the start to less than $380^{th}$ iteration, and from the start to around $5000^{th}$ iteration. Especially in the beginning of the optimization process, the improved BFE algorithm can make the loss converge to a relatively low value which is close to the optima very quickly.

Compared with the SGD and most of other adaptive algorithms, which track previous historic gradients for the estimation of the learning rate, the improved BFE algorithm mainly focuses on the future gradients towards the forward direction.

\begin{figure}[ht]
\vskip 0.2in
\begin{center}
\centerline{\includegraphics[width=\columnwidth]{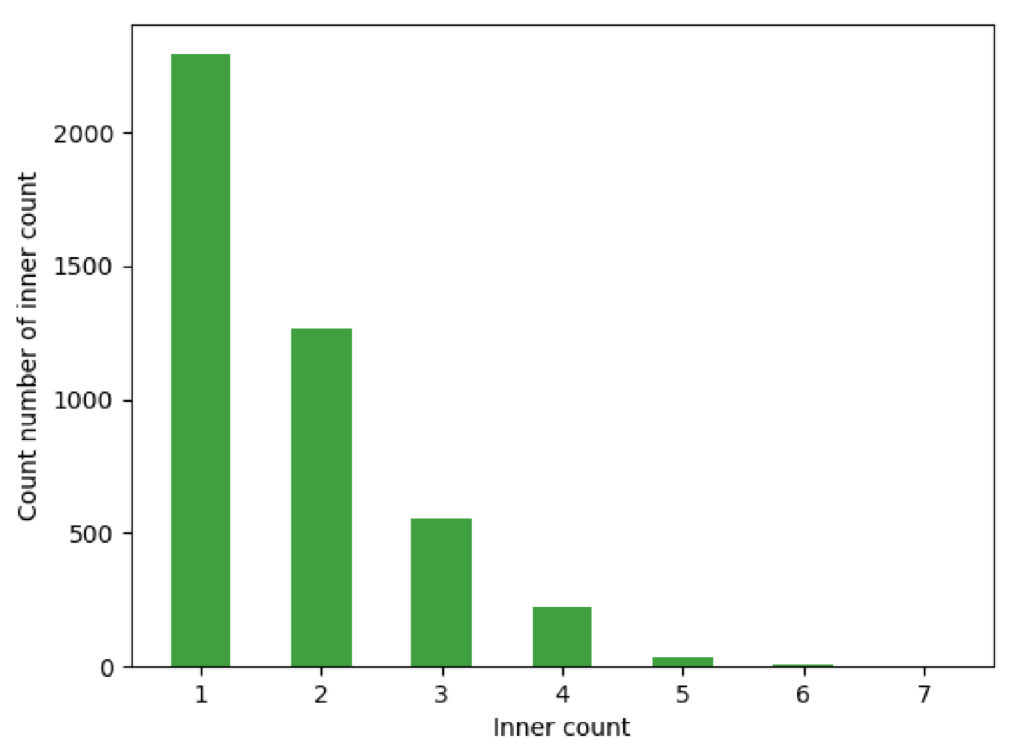}}
\caption{The histogram that shows the number of inner loops and their corresponding count number. The average number of inner loops of the improved BFE optimizer is 1.74, compared to that of the original BFE optimizer: 1.93 \cite{2022arXiv220702763C}}
\label{icml-historical2}
\end{center}
\vskip -0.2in
\end{figure}

The computational cost of the improved BFE algorithm is not significantly higher than other first-order optimization algorithm, the number of inner loops (zoom-in / zoom-out process) has been recorded during the optimization. The average inner loop number is about 1.74, compared to that of the original BFE algorithm: 1.93, both of which are less than twice for each time step’s update, as Figure 2 shows.

\begin{figure}[ht]
\vskip 0.2in
\begin{center}
\centerline{\includegraphics[width=\columnwidth]{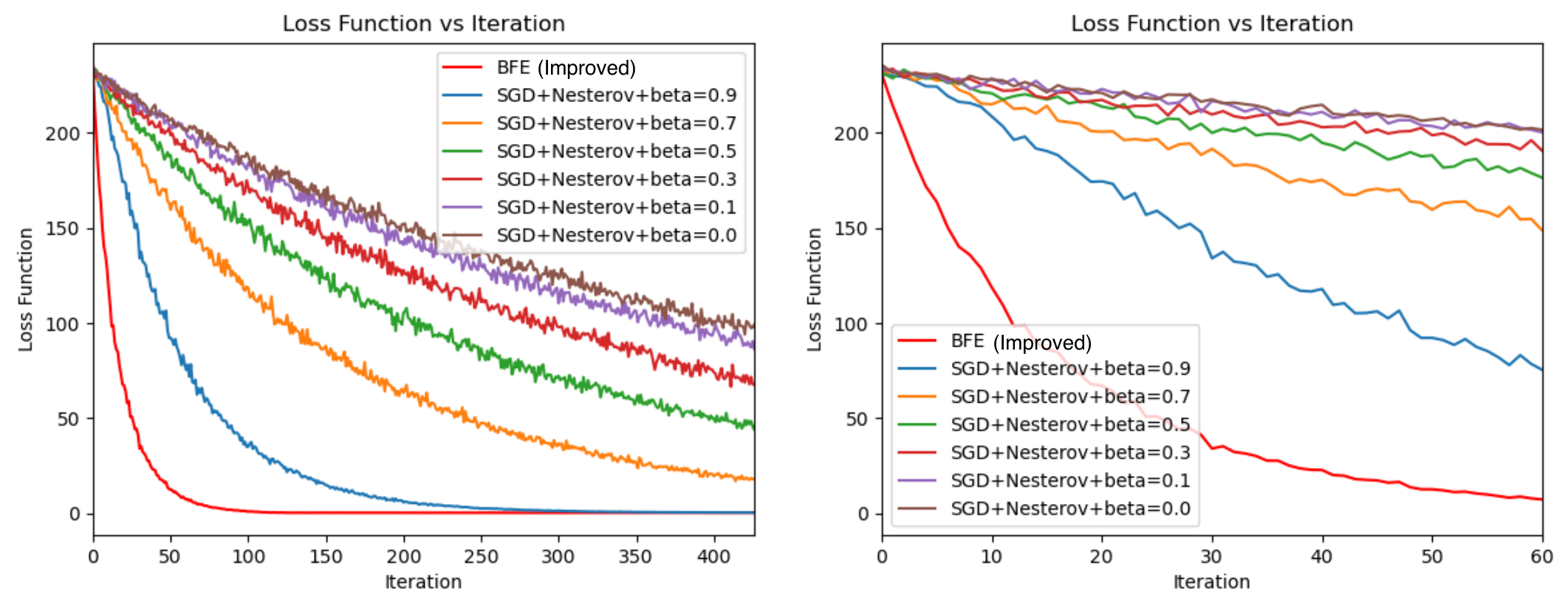}}
\caption{The loss decrease vs iterations over the whole data-set. Different colors represent the different optimization algorithms and hyper-parameters setting, in contrast to Fig. 5 in \cite{2022arXiv220702763C}}
\label{icml-historical3}
\end{center}
\vskip -0.2in
\end{figure}

As Figure 3 demonstrated, we compared the loss decrease of the BFE algorithm with those of SGD with Nesterov, each of which has different beta value from 0 to 0.9.  The results reveal that the beta value increases, and the convergence is approached faster. We also zoomed in the result between the beginning to the $60^{th}$ iteration, BFE enables the loss steeply decrease, which is expected to slide along ambient gradient of the loss landscape. The results from SGD with Nesterov did not capture the initial change of the gradient as well as BFE, because of lack of the second-order information of the loss landscape in the forward direction. This is also the reason why the warm-up strategy is necessary for manually scheduling the learning rate . A different local minimum but not optima might be approached without the warm-up under a certain of circumstances. Instead, if using BFE algorithm, there is no need to use a warm-up strategy any more due to the ability to map the second-order gradient information for forward steps. And a selection of initial learning rate can be skipped since the BFE-related algorithms can automatically obtain an optimal initial learning rate.

The loss function in this study is used as a quadratic format, which is simple but efficient to observe the convergence pattern and satisfies the characteristics of a local optima \cite{sutskever2013importance, o2015adaptive, lucas2018aggregated}. A higher order loss function in multi-parameter space with complex datasets such as Mini-ImageNet \cite{vinyals2016matching}, CIFAR-10, CIFAR-100 \cite{krizhevsky2009learning} and MNIST \cite{lecun1998gradient} will be studied and discussed in future works, which is beyond the scope of this work though.

\section{Adaptive Extension of Improved BFE Algorithm}

As the similar discussion at \cite{2022arXiv220702763C}. Besides the investigation of loss change via shrinking and amplifying learning rate by a binary scale, we can also trace gradient variation of the loss function instead of the loss change. The two methods both detect the loss landscape in the forward direction, and equivalently probe the second-order loss information without losing first-order algorithms' computational efficiency. The improved version of Adaptive BFE method was an adaptive extension to the improved BFE algorithm via adjusting learning rates per parameter via tracking gradient variation of the loss function in each parameter dimension.

\subsection{Improved BFE of gradient change}

\begin{algorithm}[H]
   \caption{Improved BFE of gradient change, the proposed algorithm in non-adaptive learning rate automation for stochastic optimization. More details are described in the text. The Loss Function can be written as a function w.r.t parameters, e.g. $Loss=f(\theta)$. Default setting for error limit ratio for Binary Detection Learning Rate is $\epsilon = 0.001$, indicating one thousandth.}
   \label{alg:example3}
\begin{algorithmic}
   
   \State Initialize learning rate $\eta$ (e.g. 0.001)
   \State Initialize $\epsilon_v$ (e.g. 0.001)
   \State Initialize $\epsilon_c > \epsilon_v$
   \State Initialize parameter vector $\theta_0$
   \State Initialize time-step $t=0$
   \While{$\theta_t$ not converged}
     \State $t=t+1$
     \If{$\epsilon_c \geq \epsilon_v$}
       \While{$\epsilon_c \geq \epsilon_v$}
         \State $\eta = \frac{\eta}{2}$
         \State $g_i = \frac{\partial f(\theta)}{\partial \theta_{t,i}}$
         \State $\theta^{*}_{t,i} = \theta_{t,i} - \eta \cdot g_i$
         \State $g^{*}_i = \frac{\partial f(\theta)}{\partial \theta^{*}_{t,i}} $
         \State $\epsilon_i = arctan(abs((g^*_i-g_i)/(1+g^*_i \cdot g_i)))$
         \State $\epsilon_c = $ function of $\epsilon_i$, \ e.g. $max(\epsilon_i)$
         \State $\epsilon_v= 1$ or any other values or functions
       \EndWhile
       \State $\theta_{t,i} = \theta^{*}_{t,i}$
     \Else
       \While{$\epsilon_c < \epsilon_v$}
         \State $\eta = 2\eta$
         \State $g_i = \frac{\partial f(\theta)}{\partial \theta_{t,i}}$
         \State $\theta^{*}_{t,i} = \theta_{t,i} - \eta \cdot g_i$

    \algstore{}
\end{algorithmic}
\end{algorithm}

\begin{algorithm}[H]
%\ContinuedFloat
\begin{algorithmic}
       \algrestore{}
       \State $g^{*}_i = \frac{\partial f(\theta)}{\partial \theta^{*}_{t,i}} $
       \State $\epsilon_i = arctan(abs((g^*_i-g_i)/(1+g^*_i \cdot g_i)))$
       \State $\epsilon_c = $ function of $\epsilon_i$, \ e.g. $max(\epsilon_i)$
       \State $\epsilon_v= 1$ or any other values or functions
       \EndWhile
       
       \State $\eta = \frac{\eta}{2}$
       \State $\theta_{t,i} = \theta_{t,i} - \eta \cdot g_i$
     \EndIf
   \State (Procedure updating $\epsilon_c$ and $\epsilon_v$, e.g. as below)
   \State $g_i = \frac{\partial f(\theta)}{\partial \theta_{t,i}}$
   \State $\theta^{*}_{t,i} = \theta_{t,i} - \eta \cdot g_i$
   \State $g^{*}_i = \frac{\partial f(\theta)}{\partial \theta^{*}_{t,i}} $
   \State $\epsilon_i = arctan(abs((g^*_i-g_i)/(1+g^*_i \cdot g_i)))$
   \State $\epsilon_c = max(\epsilon_i)$ or other functions
   \State $\epsilon_v= 1$ or any other values or functions
 \EndWhile \\
 \Return $\theta_{t}$ (Resulting Optimized Parameters)

\end{algorithmic}
\end{algorithm}

Before introducing the adaptive algorithm, the improved BFE of gradient change algorithm is described as Algorithm 3. The shrink and amplification of the learning rate can be moved to the beginning of every iteration of the zoom-in or zoom-out process, such that the updates of parameters can be further simplified. Furthermore, the $\epsilon_{comp}$ and $\epsilon_{val}$ are updated again after the completion of each time-step’s iterations. The whole optimization process is not terminated until a local optima is approached. In such a non-adaptive algorithm, all of the parameters share the same learning rate during the optimization process.

BFE of gradient change described above can be naturally extended to the corresponding adaptive version. Rather than one global learning rate, each parameter has their own learning rate $\eta_i$, which is updated via comparing $\epsilon_{c,i}$ and $\epsilon_{v,i}$ in specific dimension. Please note that, the process of updating $\epsilon_{c,i}$ and $\epsilon_{v,i}$ is shown as Procedure 5.

However, the process to update $\epsilon_{c,i}$ and $\epsilon_{v,i}$ as above are not limited to the process as Procedure 5 described, and is possible to be other forms as well if they can efficiently update these two factors. 

As Figure 4 reveals, we compared the loss curves of BFE with BFE of gradient change, the adaptive BFE of gradient change, the SGD and Adam algorithm based on univariate linear regression model (The batch size for all optimization algorithms is set to be 512). By using the similar way to \cite{2022arXiv220702763C}, Adam algorithm usually enable to make the loss decrease faster than SGD in the high-dimensional data, but in the low-dimensional (e.g. 1D or 2D) regression, loss with SGD might decrease faster \cite{gitman2018convergence}.

Furthermore, the results demonstrate the BFE of gradient change (default setting, e.g. $\epsilon_v$ = 1 degree) converges slower than the BFE, but faster than others, and the adaptive BFE version enables the loss decrease speed between SGD and Adam from the start to the $100^{th}$ iteration, which is different from \cite{2022arXiv220702763C}. The left, middle and right columns respectively demonstrate different time-step ranges in Figure 4. In the figure, each time-step means each iteration or the time-step of a batch. Studying the patterns of different versions of BFE algorithms can help us deeply understand how the multi-dimensional distribution in the parameter space affect the best-fit learning rate.

\begin{algorithm}[H]
   \caption{Improved Adaptive BFE of gradient change, the proposed algorithm in non-adaptive learning rate automation for stochastic optimization. More details are described in the text. The Loss Function can be written as a function w.r.t parameters, e.g. $Loss=f(\theta)$. Default setting for error limit ratio for Binary Detection Learning Rate is $\epsilon = 0.001$, indicating one thousandth.}
   \label{alg:example4}
\begin{algorithmic}
   
   \State Initialize learning rate $\eta_i$ (e.g. 0.001)
   \State Initialize $\epsilon_{v,i}$ (e.g. 0.001)
   \State Initialize $\epsilon_{c,i} > \epsilon_{v,i}$
   \State Initialize parameter vector $\theta_0$
   \State Initialize time-step $t=0$
   \While{$\theta_{t,i}$ not converged}
     \State $t=t+1$
     \If{$\epsilon_{c,i} \geq \epsilon_{v,i}$}
       \While{$\epsilon_{c,i} \geq \epsilon_{v,i}$}
         \State $\eta_i = \frac{\eta_i}{2}$
         \State $g_i = \frac{\partial f(\theta)}{\partial \theta_{t,i}}$
         \State $\theta^{*}_{t,i} = \theta_{t,i} - \eta \cdot g_i$
         \State $g^{*}_i = \frac{\partial f(\theta)}{\partial \theta^{*}_{t,i}} $
         \State $\epsilon_{c,i} = arctan(abs((g^*_i-g_i)/(1+g^*_i \cdot g_i)))$
         \State $\epsilon_{v,i}= 1$ or any other values or functions
       \EndWhile
       \State $\theta_{t,i} = \theta^{*}_{t,i}$
     \Else
       \While{$\epsilon_{c,i} < \epsilon_{v,i}$}
         \State $\eta_i = 2\eta_i$
         \State $g_i = \frac{\partial f(\theta)}{\partial \theta_{t,i}}$
         \State $\theta^{*}_{t,i} = \theta_{t,i} - \eta \cdot g_i$
         \State $g^{*}_i = \frac{\partial f(\theta)}{\partial \theta^{*}_{t,i}} $
         \State $\epsilon_{c,i} = arctan(abs((g^*_i-g_i)/(1+g^*_i \cdot g_i)))$
         \State $\epsilon_{v,i}= 1$ or any other values or functions
       \EndWhile
       \State $\eta_i = \frac{\eta_i}{2}$
       \State $\theta_{t,i} = \theta_{t,i} - \eta \cdot g_i$
     \EndIf
     \State (Procedure updating $\epsilon_c$ and $\epsilon_v$, e.g. as below)
     \State $g_i = \frac{\partial f(\theta)}{\partial \theta_{t,i}}$
     \State $\theta^{*}_{t,i} = \theta_{t,i} - \eta \cdot g_i$

       \algstore{}
\end{algorithmic}
\end{algorithm}

\begin{algorithm}[H]
%\ContinuedFloat
\begin{algorithmic}
       \algrestore{}
       \State $g^{*}_i = \frac{\partial f(\theta)}{\partial \theta^{*}_{t,i}} $
       \State $\epsilon_{c,i} = arctan(abs((g^*_i-g_i)/(1+g^*_i \cdot g_i)))$
       \State $\epsilon_{v,i}= 1$ or any other values or functions
       \EndWhile \\
       \Return $\theta_{t}$ per dimension (Resulting Optimized
       \State \ \ \ \ \ \ \ \ \  Parameters)
       
\end{algorithmic}
\end{algorithm}

Algorithm 4 shows the improved adaptive BFE of gradient change algorithm, which is modified from the original adaptive BFE of gradient change version. The improved version added the updating process within every outer loop in the algorithm, the updating process of which is shown in Procedure 5.

\begin{megaalgorithm}[H]
\caption{Updating $\epsilon_{c,i}$ and $\epsilon_{v,i}$ with every outer loop}
\label{alg:example5}
\begin{algorithmic}
     \State A Procedure to update $\epsilon_{c,i}$ and $\epsilon_{v,i}$ could be but not limited to the lines below:
     \State $g_i = \frac{\partial f(\theta)}{\partial \theta_{t,i}}$
     \State $\theta^{*}_{t,i} = \theta_{t,i} - \eta \cdot g_i$
     \State $g^{*}_i = \frac{\partial f(\theta)}{\partial \theta^{*}_{t,i}} $
     \State $\epsilon_{c,i} = arctan(abs((g^*_i-g_i)/(1+g^*_i \cdot g_i)))$
     \State $\epsilon_{v,i}= 1$ or any other values or functions
\end{algorithmic}
\end{megaalgorithm}

\begin{figure}[ht]
\vskip 0.2in
\begin{center}
\centerline{\includegraphics[width=\columnwidth]{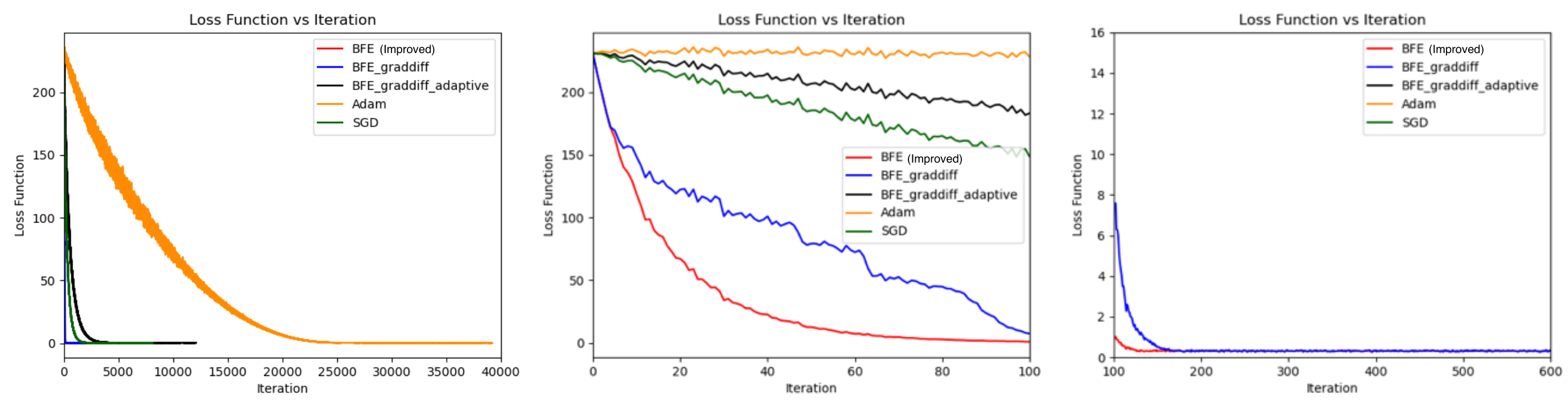}}
\caption{The loss change with the increasing iterations over the data-set. The results compare the BFE, BFE of gradient change with non-adaptive and BFE of gradient change with adaptive algorithms.}
\label{icml-historical4}
\end{center}
\vskip -0.2in
\end{figure}

\begin{figure}[ht]
\vskip 0.2in
\begin{center}
\centerline{\includegraphics[width=\columnwidth]{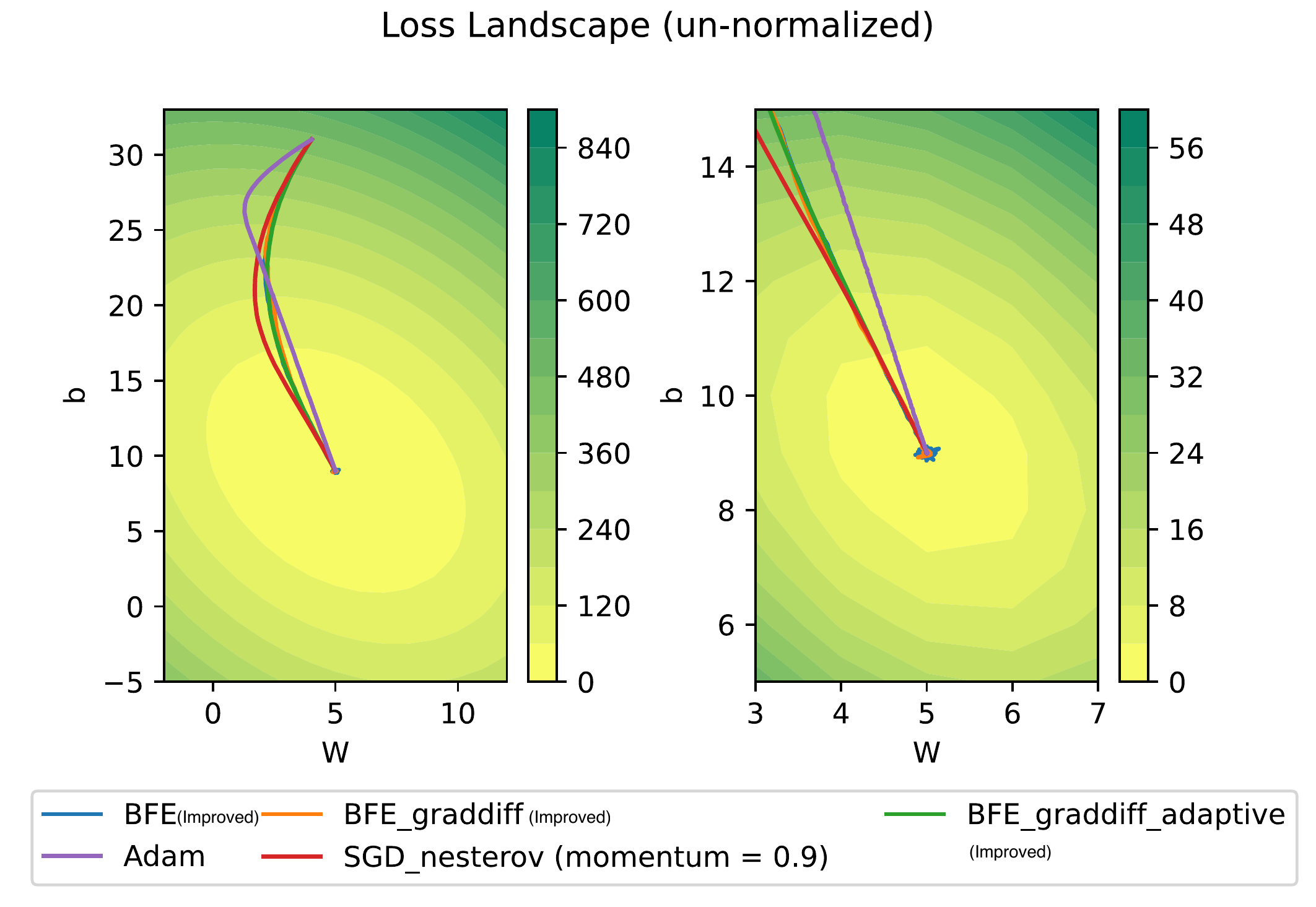}}
\caption{The comparison of BFE, the different variants of BFE, and Adam and SGD with nesterov (momentum=0.9) on the 2D loss landscape.}
\label{icml-historical5}
\end{center}
\vskip -0.2in
\end{figure}

Figure 5 reveals the loss landscape for the comparison of optimization trajectories between improved BFE, improved BFE gradient change, improved adaptive BFE gradient change algorithms and Adam and SGD with Nesterov algorithms on the parameter space. The results reveal that the improved BFE-related algorithms enable much shorter and faster trajectories to approach the optimal or minimum location, compared to the Adam and SGD with Nesterov algorithms. The excellent performance of the improved BFE-related algorithms seems to combine the advantages of both of Stochastic Gradient Descent and Adaptive-Gradient-Descent-based (e.g. Adam) algorithms, which perform a superiority in the aspects of all of the convergence speed, efficiency and accuracy during the optimization process.

\begin{figure}[ht]
\vskip 0.2in
\begin{center}
\centerline{\includegraphics[width=\columnwidth]{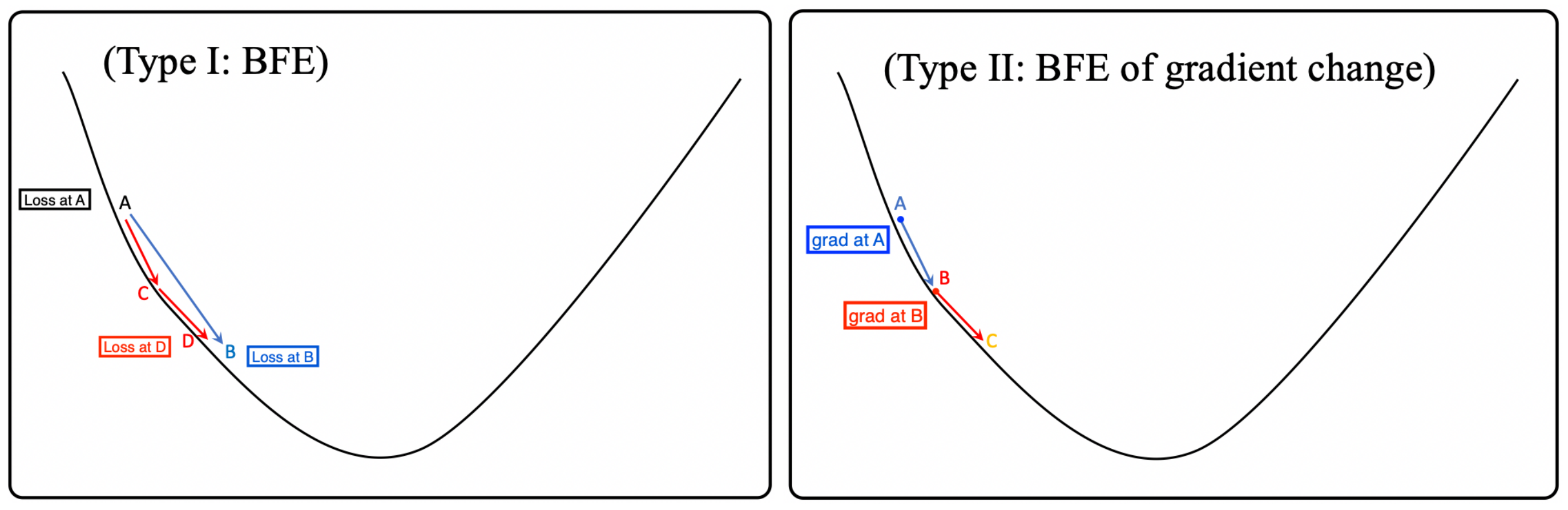}}
\caption{The brief sketch of BFE and BFE of gradient change (modified based on the figures at \cite{2022arXiv220702763C}).}
\label{icml-historical6}
\end{center}
\vskip -0.2in
\end{figure}

Figure 6 displays the sketch of a single or unit step of (improved) BFE and BFE of gradient change algorithm. The left sub-figure illustrates how to respectively obtain the loss at B and the loss at D, and then calculate their difference. The  right-figure illustrates how to respectively obtain the loss at A and the loss at B, and then compare their difference to see if it is small enough.

\begin{figure}[ht]
\vskip 0.2in
\begin{center}
\centerline{\includegraphics[width=\columnwidth]{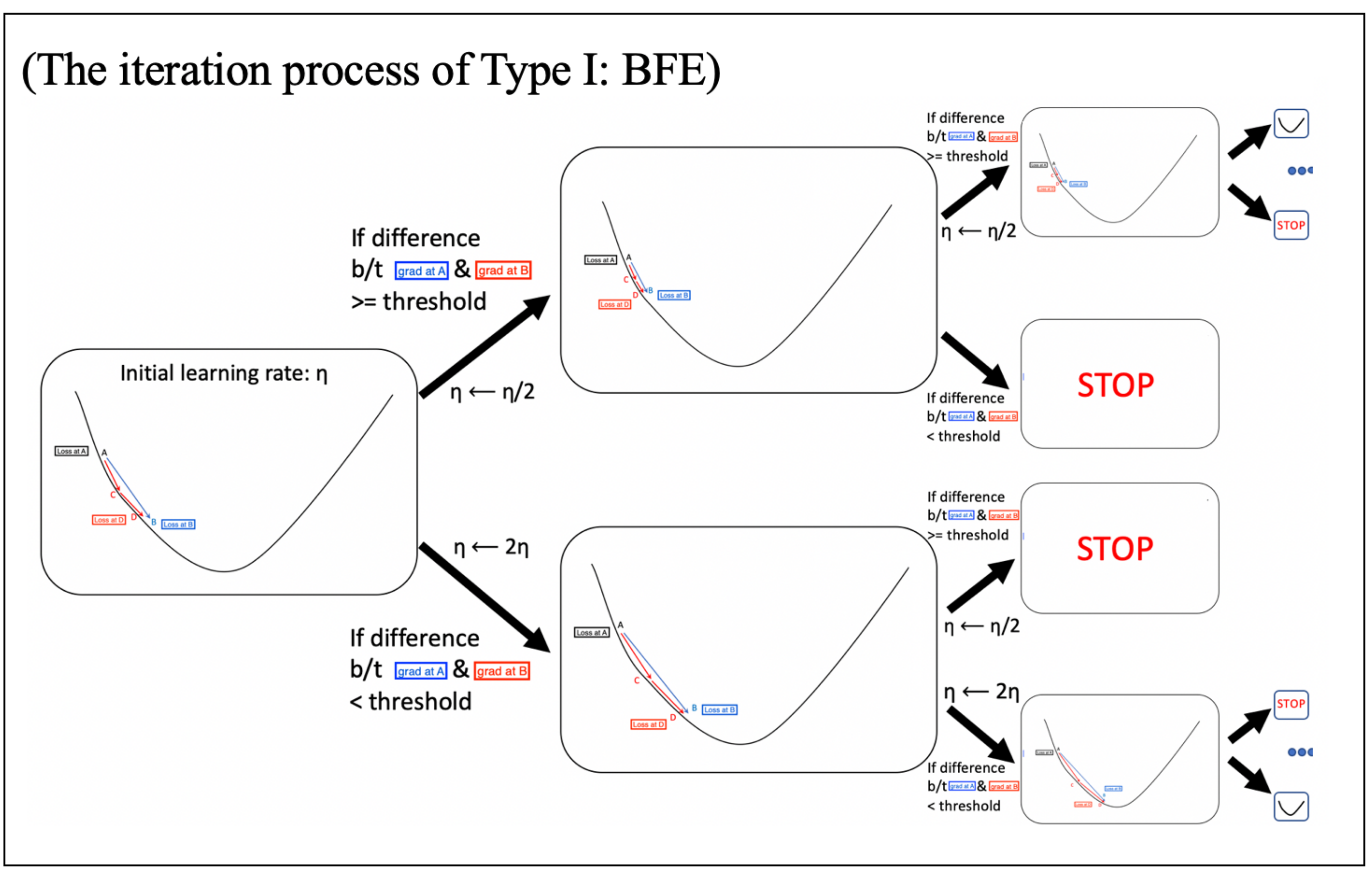}}
\caption{The brief sketch of updating process of BFE (modified based on the figures at \cite{2022arXiv220702763C}).}
\label{icml-historical7}
\end{center}
\vskip -0.2in
\end{figure}

Figure 7 demonstrates the updating process of BFE algorithm (indicated as Type I in Figure 6), where how the repeated iterative calculation of the learning rate or step size is presented. 

\begin{figure}[ht]
\vskip 0.2in
\begin{center}
\centerline{\includegraphics[width=\columnwidth]{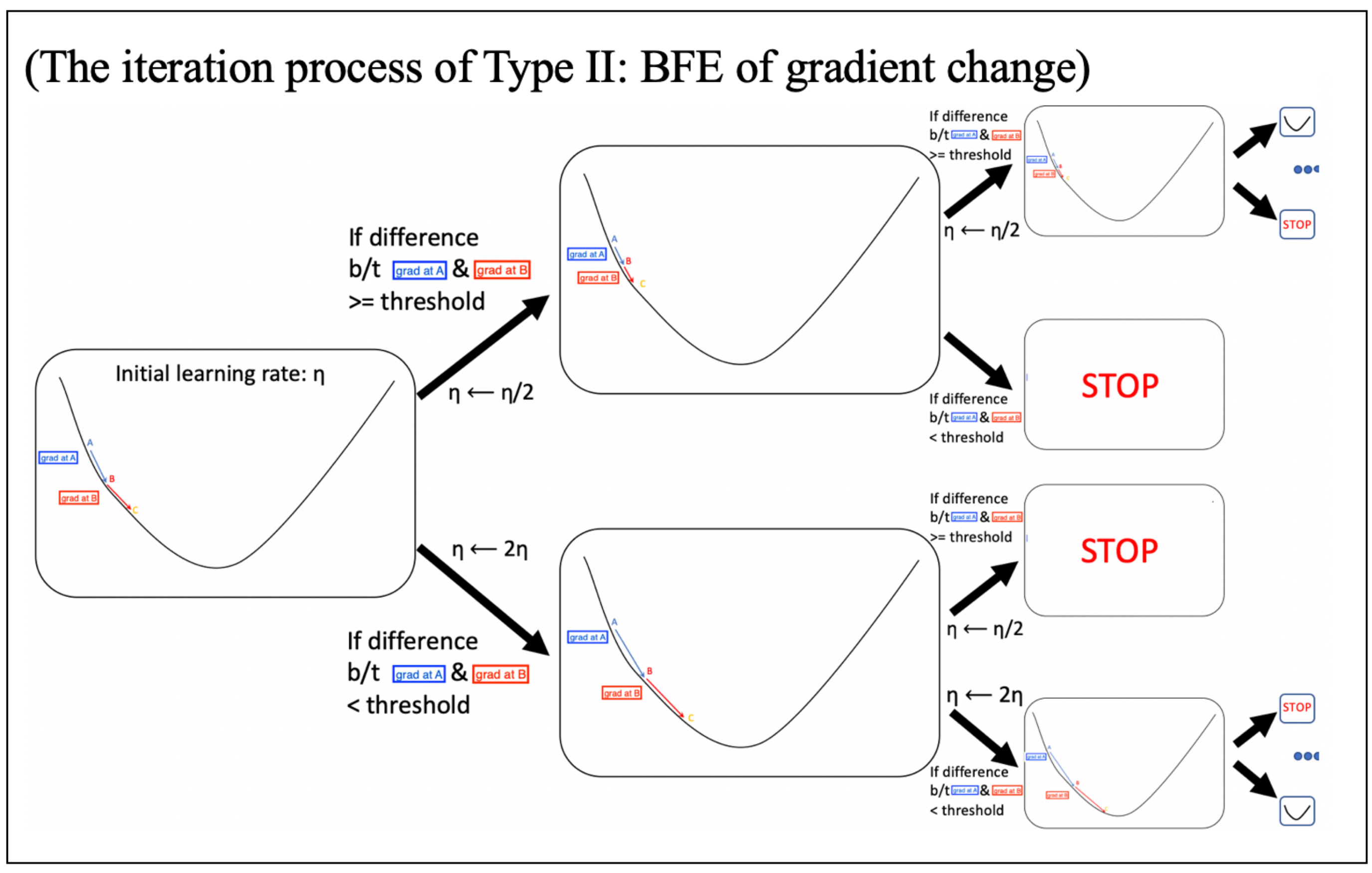}}
\caption{The brief sketch of updating process of BFE of gradient change (modified based on the figures at \cite{2022arXiv220702763C}).}
\label{icml-historical8}
\end{center}
\vskip -0.2in
\end{figure}

Similarly, Figure 8 reveals the corresponding updating process of BFE of gradient change algorithm (indicated as Type II in Figure 6). The main difference between BFE and BFE algorithms is the single unit updating step using the comparison of losses or that of sequential gradients.

\section{Summary}

In conclusion, a set of improved algorithms for learning rate automation based on Binary Forward Exploration has been proposed. This updated approach improved the parameter updating process through mapping loss landscape in the forward direction. Using the improved methods, we can more efficiently determine the learning rate during the optimization process of data-driven models. The improved algorithms show advantages such as the faster convergence speed during the beginning of optimization process, which has potential to replace warm-up strategy under some conditions. The proposed approach supplies a new viewpoint to investigate the stochastic optimization process such that we can study the linkages between the loss landscape and learning rate. In addition, the combination of fast computational efficiency based on the first-order methods with shorter convergence iterations from the second-order optimization approach. Furthermore, The improved BFE-related algorithms can enable the optimization not to use the warm-up strategy. These improved approaches are expected to play important roles in the learning rate optimization area. In additional to optimize parameters for machine learning models, it is possible for the improved BFE algorithms to be applied to the time-dependent differential problems \cite{cao2013multifluid, cao2014seasonal}, and \cite{cao20153d, cao20163d}, and \cite{Cao2017, cao2017diurnal, cao2018magnetosphere, cao2018diurnal}, and \cite{cao2019interaction, Cao2020a, Cao2020b}, and \cite{cao2020influence, cao2021using, cao2021asymmetric}. Using the similar transferring way, it is also possible to apply more numerical methods to the machine learning optimization.


\bibliography{main}

\begin{thebibliography}{33}
\providecommand{\natexlab}[1]{#1}
\providecommand{\url}[1]{\texttt{#1}}
\expandafter\ifx\csname urlstyle\endcsname\relax
  \providecommand{\doi}[1]{doi: #1}\else
  \providecommand{\doi}{doi: \begingroup \urlstyle{rm}\Url}\fi

\bibitem[{Cao}(2022)]{2022arXiv220702763C}
{Cao}, X.
\newblock {BFE and AdaBFE: A New Approach in Learning Rate Automation for
  Stochastic Optimization}.
\newblock \emph{arXiv e-prints}, art. arXiv:2207.02763, July 2022.

\bibitem[Cao \& Paty(2013)Cao and Paty]{cao2013multifluid}
Cao, X. and Paty, C.
\newblock Multifluid mhd simulation of the magnetosphere of uranus.
\newblock In \emph{AGU Fall Meeting Abstracts}, volume 2013, pp.\  SM21A--2154,
  2013.

\bibitem[Cao \& Paty(2017{\natexlab{a}})Cao and Paty]{Cao2017}
Cao, X. and Paty, C.
\newblock Diurnal and seasonal variability of uranus's magnetosphere.
\newblock \emph{Journal of Geophysical Research: Space Physics}, 122:\penalty0
  6318--6331, 2017{\natexlab{a}}.
\newblock ISSN 21699402.
\newblock \doi{10.1002/2017JA024063}.

\bibitem[Cao \& Paty(2017{\natexlab{b}})Cao and Paty]{cao2017diurnal}
Cao, X. and Paty, C.
\newblock Diurnal and seasonal variability of uranus' magnetopause under
  different imf.
\newblock In \emph{AGU Fall Meeting Abstracts}, volume 2017, pp.\  P34C--04,
  2017{\natexlab{b}}.

\bibitem[Cao \& Paty(2018{\natexlab{a}})Cao and Paty]{cao2018magnetosphere}
Cao, X. and Paty, C.
\newblock The magnetosphere of uranus.
\newblock In \emph{Oxford Research Encyclopedia of Planetary Science}.
  2018{\natexlab{a}}.

\bibitem[Cao \& Paty(2021)Cao and Paty]{cao2021asymmetric}
Cao, X. and Paty, C.
\newblock Asymmetric structure of uranus' magnetopause controlled by imf and
  planetary rotation.
\newblock \emph{Geophysical Research Letters}, 48\penalty0 (4):\penalty0
  e2020GL091273, 2021.

\bibitem[Cao \& Paty(2014)Cao and Paty]{cao2014seasonal}
Cao, X. and Paty, C.~S.
\newblock A seasonal study of uranus' magnetosphere.
\newblock In \emph{AGU Fall Meeting Abstracts}, volume 2014, pp.\  SM51E--4283,
  2014.

\bibitem[Cao \& Paty(2015)Cao and Paty]{cao20153d}
Cao, X. and Paty, C.~S.
\newblock 3d multifluid mhd simulation for uranus and neptune: the seasonal
  variations of their magnetosphere.
\newblock In \emph{2015 AGU Fall Meeting}. AGU, 2015.

\bibitem[Cao \& Paty(2016)Cao and Paty]{cao20163d}
Cao, X. and Paty, C.~S.
\newblock 3d multifluid mhd simulations at uranus and neptune: Seasonal
  variations of their magnetospheres.
\newblock In \emph{AGU Fall Meeting Abstracts}, pp.\  SM44B--08, 2016.

\bibitem[Cao \& Paty(2018{\natexlab{b}})Cao and Paty]{cao2018diurnal}
Cao, X. and Paty, C.~S.
\newblock Diurnal and seasonal variability of uranus' and neptune's
  magnetopause under different imf.
\newblock In \emph{AGU Fall Meeting Abstracts}, volume 2018, pp.\  P33E--3879,
  2018{\natexlab{b}}.

\bibitem[Cao et~al.(2013)Cao, Lu, Yang, Liu, Yang, and
  Zhao]{Cao2013Trajectorymethodof3Dtestparticles}
Cao, X., Lu, J., Yang, Z., Liu, Z., Yang, Y., and Zhao, M.
\newblock Trajectory method of 3{D} test particles in global transport in
  magnetosphere.
\newblock \emph{Journal of Space Science}, 33\penalty0 (3):\penalty0 240--249,
  2013.

\bibitem[Cao et~al.(2019)Cao, Halekas, McFadden, and
  Glassmeier]{cao2019interaction}
Cao, X., Halekas, J.~S., McFadden, J.~P., and Glassmeier, K.-H.
\newblock The interaction of lunar ions with the ambient environment in the
  terrestrial magnetospheric tail lobe.
\newblock In \emph{AGU Fall Meeting Abstracts}, volume 2019, pp.\  SM33F--3283,
  2019.

\bibitem[Cao et~al.(2020{\natexlab{a}})Cao, Halekas, Poppe, Chu, and
  Glassmeier]{Cao2020a}
Cao, X., Halekas, J., Poppe, A., Chu, F., and Glassmeier, K.~H.
\newblock The acceleration of lunar ions by magnetic forces in the terrestrial
  magnetotail lobes.
\newblock \emph{Journal of Geophysical Research: Space Physics}, 125:\penalty0
  1--12, 2020{\natexlab{a}}.
\newblock ISSN 21699402.
\newblock \doi{10.1029/2020JA027829}.

\bibitem[Cao et~al.(2020{\natexlab{b}})Cao, Halekas, Chu, Kistler, Poppe, and
  Glassmeier]{Cao2020b}
Cao, X., Halekas, J.~S., Chu, F., Kistler, M., Poppe, A.~R., and Glassmeier, K.
\newblock Plasma convection in the terrestrial magnetotail lobes measured near
  the moon's orbit.
\newblock \emph{Geophysical Research Letters}, 47:\penalty0 1--7,
  2020{\natexlab{b}}.
\newblock ISSN 0094-8276.
\newblock \doi{10.1029/2020gl090217}.

\bibitem[Cao et~al.(2020{\natexlab{c}})Cao, Halekas, Chu, Kistler, Poppe, and
  Glassmeier]{cao2020influence}
Cao, X., Halekas, J.~S., Chu, F., Kistler, M., Poppe, A.~R., and Glassmeier,
  K.-H.
\newblock The influence of the upstream conditions on the plasma convection in
  the distant tail lobes.
\newblock In \emph{AGU Fall Meeting Abstracts}, volume 2020, pp.\  SM055--0001,
  2020{\natexlab{c}}.

\bibitem[Cao et~al.(2021)Cao, Halekas, Haaland, Ruhunusiri, Poppe, and
  Glassmeier]{cao2021using}
Cao, X., Halekas, J.~S., Haaland, S., Ruhunusiri, S., Poppe, A.~R., and
  Glassmeier, K.-H.
\newblock Using machine learning to characterize solar wind driving of
  convection in the terrestrial magnetotail lobes.
\newblock In \emph{AGU Fall Meeting 2021}. AGU, 2021.

\bibitem[Cao et~al.(2022)Cao, Halekas, Haaland, Ruhunusiri, and
  Glassmeier]{cao2022machine}
Cao, X., Halekas, J.~S., Haaland, S., Ruhunusiri, S., and Glassmeier, K.-H.
\newblock Machine learning solar wind driving magnetospheric convection in tail
  lobes.
\newblock \emph{arXiv preprint arXiv:2202.01383}, 2022.

\bibitem[Chu et~al.(2021{\natexlab{a}})Chu, Girazian, Duru, Ramstad, Halekas,
  Gurnett, Cao, and Kopf]{chu2021dayside}
Chu, F., Girazian, Z., Duru, F., Ramstad, R., Halekas, J., Gurnett, D., Cao,
  X., and Kopf, A.
\newblock The dayside ionopause of mars: Solar wind interaction, pressure
  balance, and comparisons with venus.
\newblock \emph{Journal of Geophysical Research: Planets}, 126\penalty0
  (11):\penalty0 e2021JE006936, 2021{\natexlab{a}}.

\bibitem[Chu et~al.(2021{\natexlab{b}})Chu, Halekas, Cao, Mcfadden, Bonnell,
  and Glassmeier]{chu2021electrostatic}
Chu, F., Halekas, J.~S., Cao, X., Mcfadden, J.~P., Bonnell, J.~W., and
  Glassmeier, K.-H.
\newblock Electrostatic waves and electron heating observed over lunar crustal
  magnetic anomalies.
\newblock \emph{Journal of Geophysical Research: Space Physics}, 126\penalty0
  (4):\penalty0 e2020JA028880, 2021{\natexlab{b}}.

\bibitem[Duchi et~al.(2011)Duchi, Hazan, and Singer]{duchi2011adaptive}
Duchi, J., Hazan, E., and Singer, Y.
\newblock Adaptive subgradient methods for online learning and stochastic
  optimization.
\newblock \emph{Journal of machine learning research}, 12\penalty0 (7), 2011.

\bibitem[Gitman et~al.(2018)Gitman, Dilipkumar, and
  Parr]{gitman2018convergence}
Gitman, I., Dilipkumar, D., and Parr, B.
\newblock Convergence analysis of gradient descent algorithms with proportional
  updates.
\newblock \emph{arXiv preprint arXiv:1801.03137}, 2018.

\bibitem[Kingma \& Ba(2014)Kingma and Ba]{kingma2014adam}
Kingma, D.~P. and Ba, J.
\newblock Adam: A method for stochastic optimization.
\newblock \emph{arXiv preprint arXiv:1412.6980}, 2014.

\bibitem[Krizhevsky et~al.(2009)Krizhevsky, Hinton,
  et~al.]{krizhevsky2009learning}
Krizhevsky, A., Hinton, G., et~al.
\newblock Learning multiple layers of features from tiny images.
\newblock 2009.

\bibitem[LeCun et~al.(1998)LeCun, Bottou, Bengio, and
  Haffner]{lecun1998gradient}
LeCun, Y., Bottou, L., Bengio, Y., and Haffner, P.
\newblock Gradient-based learning applied to document recognition.
\newblock \emph{Proceedings of the IEEE}, 86\penalty0 (11):\penalty0
  2278--2324, 1998.

\bibitem[Liu et~al.(2012)Liu, Lu, Kabin, Yang, Zhao, and Cao]{liu2012dipole}
Liu, Z.-Q., Lu, J., Kabin, K., Yang, Y., Zhao, M., and Cao, X.
\newblock Dipole tilt control of the magnetopause for southward imf from global
  magnetohydrodynamic simulations.
\newblock \emph{Journal of Geophysical Research: Space Physics}, 117\penalty0
  (A7), 2012.

\bibitem[Liuzzo et~al.(2021)Liuzzo, Poppe, Halekas, Simon, and
  Cao]{liuzzo2021investigating}
Liuzzo, L., Poppe, A.~R., Halekas, J.~S., Simon, S., and Cao, X.
\newblock Investigating the moon's interaction with the terrestrial magnetotail
  lobe plasma.
\newblock \emph{Geophysical Research Letters}, 48\penalty0 (9):\penalty0
  e2021GL093566, 2021.

\bibitem[Lucas et~al.(2018)Lucas, Sun, Zemel, and Grosse]{lucas2018aggregated}
Lucas, J., Sun, S., Zemel, R., and Grosse, R.
\newblock Aggregated momentum: Stability through passive damping.
\newblock \emph{arXiv preprint arXiv:1804.00325}, 2018.

\bibitem[Nesterov(1983)]{nesterov1983method}
Nesterov, Y.~E.
\newblock A method for solving the convex programming problem with convergence
  rate o (1/k\^{} 2).
\newblock In \emph{Dokl. akad. nauk Sssr}, volume 269, pp.\  543--547, 1983.

\bibitem[O’donoghue \& Candes(2015)O’donoghue and Candes]{o2015adaptive}
O’donoghue, B. and Candes, E.
\newblock Adaptive restart for accelerated gradient schemes.
\newblock \emph{Foundations of computational mathematics}, 15\penalty0
  (3):\penalty0 715--732, 2015.

\bibitem[Sutskever et~al.(2013)Sutskever, Martens, Dahl, and
  Hinton]{sutskever2013importance}
Sutskever, I., Martens, J., Dahl, G., and Hinton, G.
\newblock On the importance of initialization and momentum in deep learning.
\newblock In \emph{International conference on machine learning}, pp.\
  1139--1147. PMLR, 2013.

\bibitem[Tieleman et~al.(2012)Tieleman, Hinton, et~al.]{tieleman2012lecture}
Tieleman, T., Hinton, G., et~al.
\newblock Lecture 6.5-rmsprop: Divide the gradient by a running average of its
  recent magnitude.
\newblock \emph{COURSERA: Neural networks for machine learning}, 4\penalty0
  (2):\penalty0 26--31, 2012.

\bibitem[Vinyals et~al.(2016)Vinyals, Blundell, Lillicrap, Wierstra,
  et~al.]{vinyals2016matching}
Vinyals, O., Blundell, C., Lillicrap, T., Wierstra, D., et~al.
\newblock Matching networks for one shot learning.
\newblock \emph{Advances in neural information processing systems}, 29, 2016.

\bibitem[Zeiler(2012)]{zeiler2012adadelta}
Zeiler, M.~D.
\newblock Adadelta: an adaptive learning rate method.
\newblock \emph{arXiv preprint arXiv:1212.5701}, 2012.

\end{thebibliography}
\bibliographystyle{icml2021}

\newpage

\begin{appendices}
\end{appendices}

\begin{algorithm}[H]
   \caption{Multiple Forward Exploration (MFE), e.g. if multiple number = 3, it is then Triple Forward Exploration version. The Loss Function can be written as a function w.r.t parameters, e.g. $Loss=f(\theta)$. Default setting for error limit ratio for Multiple Detection Learning Rate is $\epsilon = 0.001$, indicating one thousandth.}
   \label{alg:example6}
\begin{algorithmic}
   \State Initialize learning rate $\eta$ (e.g. 0.001)
   \State Initialize $\epsilon_v$ (e.g. 0.001)
   % \STATE \textcolor{orange}{Initialize} $\epsilon_v$ (e.g. 0.001)
   \State Initialize $\epsilon_c > \epsilon_v$
   \State Initialize parameter vector $\theta_0$
   \State Initialize time-step $t=0$
   \While{$\theta_t$ not converged}
     \State $t=t+1$
     \If{$\epsilon_c \geq \epsilon_v$}
       \While{$\epsilon_c \geq \epsilon_v$}
         \State $\theta^{*}_{t} = \theta_t - \eta \frac{\partial f(\theta)}{\partial \theta_{t}}$
         \State $\theta^{+}_{t} = \theta_t - \frac{\eta}{3} \frac{\partial f(\theta)}{\partial \theta_{t}}$
         \State $\theta^{'}_{t} = \theta^{+}_{t} - \frac{\eta}{3} \frac{\partial f(\theta)}{\partial \theta^{+}_{t}}$
         \State $\theta^{\#}_{t} = \theta^{'}_{t} - \frac{\eta}{3} \frac{\partial f(\theta)}{\partial \theta^{'}_{t}}$
         
         \State $Loss1 = [f(\theta)]_{\theta^{*}_{t}}$ $ $ $ $ (loss value at $\theta^{*}_{t}$)
         \State $Loss2 = [f(\theta)]_{\theta^{\#}_{t}}$ $ $ $ $ (loss value at $\theta^{\#}_{t}$)
         \State $\epsilon_c = |Loss2-Loss1|$
         \State $\epsilon_v= 0.5 \cdot (|Loss2|+|Loss1|) \cdot \epsilon$
         \State \ \ \ \ \ \ \ \ \  or $min(|Loss2|\cdot \epsilon, |Loss1|\cdot \epsilon)$
         \State \ \ \ \ \ \ \ \ \  or any other predefined factor
         \State \ \ \ \ \ \ \ \ \  or functions, e.g. decay with time-step 
         \State \ \ \ \ \ \ \ \ \  or epochs
         \State $\eta = \frac{\eta}{3}$
       \EndWhile
       \State $\eta = 3\eta$
       \State $\theta_{t} = \theta^{*}_{t}$
       \State (Procedure updating $\epsilon_c$ and $\epsilon_v$, e.g. as below)
       \State $\theta^{*}_{t} = \theta_t - \eta \frac{\partial f(\theta)}{\partial \theta_{t}}$
         \State $\theta^{+}_{t} = \theta_t - \frac{\eta}{3} \frac{\partial f(\theta)}{\partial \theta_{t}}$
         \State $\theta^{'}_{t} = \theta^{+}_{t} - \frac{\eta}{3} \frac{\partial f(\theta)}{\partial \theta^{+}_{t}}$
         \State $\theta^{\#}_{t} = \theta^{'}_{t} - \frac{\eta}{3} \frac{\partial f(\theta)}{\partial \theta^{'}_{t}}$
         \State $Loss1 = [f(\theta)]_{\theta^{*}_{t}}$ $ $ $ $ (loss value at $\theta^{*}_{t}$)
         \State $Loss2 = [f(\theta)]_{\theta^{\#}_{t}}$ $ $ $ $ (loss value at $\theta^{\#}_{t}$)
         \State $\epsilon_c = |Loss2-Loss1|$
         \State $\epsilon_v= 0.5 \cdot (|Loss2|+|Loss1|) \cdot \epsilon$
         \State \ \ \ \ \ \ \ \ \  or $min(|Loss2|\cdot \epsilon, |Loss1|\cdot \epsilon)$
         \State \ \ \ \ \ \ \ \ \  or any other predefined factor
         \State \ \ \ \ \ \ \ \ \  or functions, e.g. decay with time-step 
         \State \ \ \ \ \ \ \ \ \  or epochs

     \Else
       \While{$\epsilon_c < \epsilon_v$}
         \State $\theta^{+}_{t} = \theta_t - \eta \frac{\partial f(\theta)}{\partial \theta_{t}}$

         \algstore{}
\end{algorithmic}
\end{algorithm}

\begin{algorithm}[H]
%\ContinuedFloat
\begin{algorithmic}
       \algrestore{}
       \State $\theta^{'}_{t} = \theta^{+}_t - \eta \frac{\partial f(\theta)}{\partial \theta^{+}_{t}}$
       \State $\theta^{\#}_{t} = \theta^{'}_t - \eta \frac{\partial f(\theta)}{\partial \theta^{'}_{t}}$
       \State $\theta^{*}_{t} = \theta_{t} - 3\eta \frac{\partial f(\theta)}{\partial \theta_{t}}$
       \State $Loss1 = [f(\theta)]_{\theta^{\#}_{t}}$ $ $ $ $ (loss value at $\theta^{\#}_{t}$)
       \State $Loss2 = [f(\theta)]_{\theta^{*}_{t}}$ $ $ $ $ (loss value at $\theta^{*}_{t}$)
       \State $\epsilon_c = |Loss2-Loss1|$
       \State $\epsilon_v= 0.5 \cdot (|Loss2|+|Loss1|) \cdot \epsilon$
       \State \ \ \ \ \ \ \ \ \  or $min(|Loss2|\cdot \epsilon, |Loss1|\cdot \epsilon)$
       \State \ \ \ \ \ \ \ \ \  or any other factor or functions
         \State $\eta = 3\eta$
       \EndWhile
       %\algstore{}
       \State $\eta = \frac{\eta}{3}$
       \State $\theta_{t} = \theta^{+}_{t}$
       \State (Procedure updating $\epsilon_c$ and $\epsilon_v$, e.g. as below)
       \State $\theta^{+}_{t} = \theta_t - \eta \frac{\partial f(\theta)}{\partial \theta_{t}}$
       \State $\theta^{'}_{t} = \theta^{+}_t - \eta \frac{\partial f(\theta)}{\partial \theta^{+}_{t}}$
       \State $\theta^{\#}_{t} = \theta^{'}_t - \eta \frac{\partial f(\theta)}{\partial \theta^{'}_{t}}$
       \State $\theta^{*}_{t} = \theta_{t} - 3\eta \frac{\partial f(\theta)}{\partial \theta_{t}}$
       \State $Loss1 = [f(\theta)]_{\theta^{\#}_{t}}$ $ $ $ $ (loss value at $\theta^{\#}_{t}$)
       \State $Loss2 = [f(\theta)]_{\theta^{*}_{t}}$ $ $ $ $ (loss value at $\theta^{*}_{t}$)
       \State $\epsilon_c = |Loss2-Loss1|$
       \State $\epsilon_v= 0.5 \cdot (|Loss2|+|Loss1|) \cdot \epsilon$
       \State \ \ \ \ \ \ \ \ \  or $min(|Loss2|\cdot \epsilon, |Loss1|\cdot \epsilon)$
       \State \ \ \ \ \ \ \ \ \  or any other factor or functions

     \EndIf
   \EndWhile \\
   \Return $\theta_{t}$ (Resulting Optimized Parameters)
   
\end{algorithmic}
\end{algorithm}

\end{document}